\pdfoutput=1
\documentclass[11pt]{article}

\usepackage{CJKutf8}

\usepackage{acl}

\usepackage{times}
\usepackage{latexsym}
\usepackage{diagbox}

\usepackage[T1]{fontenc}
\usepackage[utf8]{inputenc}

\usepackage{microtype}

\usepackage{inconsolata}

\usepackage{amsmath}
\usepackage{cleveref}
\usepackage{booktabs}
\usepackage{multirow}
\usepackage{graphicx}
\usepackage{amssymb}
\usepackage{subfigure}
\usepackage{tikz}
\usepackage{scalerel}

\newcommand\cruleblacky[3][black]{\begingroup
    \fboxsep=0pt\raisebox{\fboxrule}{%
    \fbox{\textcolor{#1}{%
    \rule{\dimexpr #2-2\fboxrule\relax}{\dimexpr #3-2\fboxrule\relax}%
}}}\endgroup}

\newcommand{\yellowsign}[1][black, fill=yellow]{\scalebox{.15}{\tikz \draw[rounded corners=0.1pt,#1] (0,1) -- (1.25,3) -- (2.5,1) -- cycle;}}
\newcommand{\greysign}[1][black, fill=gray]{\scalebox{.15}{\tikz \draw[rounded corners=0.1pt,#1] (0,1) -- (1.25,3) -- (2.5,1) -- cycle;}}

\newcommand{\redsign}{\vspace{-.1cm}\cruleblacky[red]{9pt}{9pt}}

\title{Categorical Syllogisms Revisited: A Review of the Logical Reasoning Abilities of LLMs for Analyzing Categorical Syllogisms}

\author{Shi Zong, Jimmy Lin\\David R. Cheriton School of Computer Science\\University of Waterloo\\\texttt{\{s4zong, jimmylin\}@uwaterloo.ca}}

\begin{document}

\maketitle

\begin{abstract}
There has been a huge number of benchmarks proposed to evaluate how large language models (LLMs) behave for logic inference tasks. 
However, it remains an open question how to properly evaluate this ability.
In this paper, we provide a systematic overview of prior works on the logical reasoning ability of LLMs for analyzing categorical syllogisms.
We first investigate all the possible variations for categorical syllogisms from a purely logical perspective and then examine the underlying configurations (i.e., mood and figure) tested by existing datasets. 
Our results indicate that compared to template-based synthetic datasets, crowdsourcing approaches normally sacrifice the coverage of configurations (i.e., mood and figure) of categorical syllogisms for more language variations, thus bringing challenges to fully testing LLMs under different situations.
We then summarize the findings and observations for the performance of LLMs to infer the validity of syllogisms from the current literature. The error rate breakdown analyses suggest that the interpretation of quantifiers seems to be the current bottleneck that limits the performance of the LLMs and is thus worth more attention.
Finally, we discuss several points that might be worth considering when researchers plan to release categorical syllogism datasets. 
We hope our work will provide a timely review of the current literature regarding categorical syllogisms, and motivate more interdisciplinary research between communities, specifically computational linguists and logicians.
\end{abstract}

\begin{CJK*}{UTF8}{gbsn}

\section{Introduction}

Large language models (LLMs) have achieved remarkable performance on a variety of tasks \cite{NEURIPS2020_1457c0d6, NEURIPS2022_9d560961, bubeck2023sparks}.
Over the years, a large number of benchmarks have been proposed that try to evaluate the different abilities of LLMs, many of which are designed for measuring logical reasoning ability using a variety of tasks.
\citet{habernal-etal-2018-argument} propose an argument reading comprehension task to test deductive reasoning.
CLUTRR \cite{sinha2019clutrr} tests inductive reasoning capabilities by requiring to infer kinship relations between characters in short stories.
ReClor \cite{yu2020reclor}, MMLU \cite{hendrycks2021measuring}, and LogiQA \cite{ijcai2020p501} contain multiple-choice reading comprehension questions to evaluate diverse forms of logical reasoning.
Datasets such as SylloBase \cite{wu-etal-2023-hence} and FOLIO \cite{han2022folio} require LLMs to conduct inferences using syllogism logic or first-order logic.
Among these datasets, many consist of questions that are directly taken from exams. For example, MMLU \cite{hendrycks2021measuring} contains practice questions from tests such as the Graduate Record Examination (GRE), and ReClor \cite{yu2020reclor} collects problems from the Law School Admission Test (LSAT).

A fundamental question behind these datasets is: \textit{how to design a benchmark to ensure a fair and comprehensive evaluation of logic reasoning abilities?}
This question is particularly important when the test questions are self-generated, instead of directly collected from established examinations for humans mentioned above.
Problems in those human examinations are developed over decades and are designed in support of theories such as psychometrics and measurement in education.
Thus, having rigorous analyses of current benchmarks designed for LLMs would ensure that we can track the development progress of LLMs accurately.

In this work, we make progress in answering the above question for a specific task: categorical syllogisms.\footnote{Unless specified, the term ``categorical syllogisms'' is also directly written as ``syllogisms'' (due to space issues).}
Besides the reason that to the best of our knowledge, there is no prior work on analyzing categorical syllogism datasets from a designing principle's perspective, we note some other compelling reasons for choosing this task.
(1) Syllogisms are inarguably the most basic building block in logical reasoning abilities. Having a deeper understanding of syllogism inference is thus beneficial when designing models for solving more complex reasoning tasks.
(2) Categorical syllogisms have a finite number of situations (discussed in \Cref{sec:cate_syll}), which could enable a complete check of all the possible cases for LLMs.
(3) How to properly solve categorical syllogisms has been studied by logicians over decades. There is a huge literature that we can draw inspiration from to help understand how LLMs behave or make LLMs more efficient.

To sum up, our intention is not to propose new models to achieve the start-of-the-performance on certain datasets, nor introduce new benchmarks.
Rather, we hope to take a step back and systematically review all existing work to understand where we are right now.
Our goal is to check missing pieces and identify areas that are worth clarifying or need future research. 
Specifically, in this paper, we make the following contributions:
% \vspace{-.2cm}
\begin{itemize}
\itemsep -1pt
\item We investigate all existing categorical syllogism datasets in literature along with their properties in \Cref{sec:existing_datasets_overview}. A checklist covering all the variations of categorical syllogisms from a purely logician's perspective is provided and we then examine the coverage of different cases for existing benchmarks.
\item We summarize all prior findings related to the performance of LLMs for checking the validity of syllogisms in \Cref{sec:exp}. By presenting an error rate breakdown by the mood and figure of syllogisms, we highlight the importance of enhancing the abilities of LLMs for interpreting quantifiers. 
\item We provide suggestions for the future releases of categorical syllogism datasets in \Cref{sec:discussion}, including clarifying certain issues such as existential import, providing complete annotations, and building datasets containing ordinary arguments.
\end{itemize}

\section{A Concise Introduction to Syllogisms}

In this section, we provide a brief introduction to categorical syllogisms from a logician's perspective.
We will show in \Cref{sec:variations_list,sec:bottleneck} that these preparations will help us evaluate current syllogism datasets and better understand the bottleneck of the performance of LLMs.

\subsection{Categorical Syllogisms}
\label{sec:cate_syll}

\begin{table}[]
\small
\centering
\begin{tabular}{l}
Major Premise: All Greeks are humans.\\
Minor Premise: All Athenians are Greeks.\\
Conclusion: Therefore, all Athenians are humans.
\end{tabular}
\caption{An example of a standard-form categorical syllogism (mood AAA, figure 1, configuration AAA-1).}
\label{fig:example_syllogism}
\end{table}

\begin{table}[]
\centering
\resizebox{0.49\textwidth}{!}{
\begin{tabular}{llc}
\toprule
Proposition & Type & Gen. quant. \\\midrule
All S are P. & Universal Affirmative (A) & $S \subseteq P$ \\
No S is P. & Universal Negative (E) & $S \cap P=\varnothing$\\
Some S is P. & Particular Affirmative (I) & $S\cap P\neq\varnothing$ \\
Some S is not P. & Particular Negative (O) & $S- P\neq\varnothing$ \\\bottomrule
\end{tabular}
}
\caption{Types of propositions with corresponding expressions using generalized quantifier theory.}
\label{tb:type_propositions}
\end{table}

\paragraph{Categorical Propositions.}
A categorical proposition relates two classes, or categories.
In practice, we care most about a categorical proposition in its standard form, which can be written as: \texttt{Quantifier (Subject) Copula (Predicate)}.
There are only 4 kinds of standard-form categorical propositions, listed in \Cref{tb:type_propositions}.

\paragraph{Terms.} 
A syllogism contains three terms: the predicate term (P), the middle term (M), and the subject term (S).
The middle term never occurs in the conclusion but always appears in both premises.
The term that occurs as the predicate and the subject of the conclusion is called the major term and minor term, respectively.

\paragraph{Standard-Form Categorical Syllogisms.}
A categorical syllogism in its standard form must meet the following two requirements: (1) Its premises and conclusion are all standard-form categorical propositions (A, E, I, or O; see \Cref{tb:type_propositions}); and (2) Propositions are arranged in standard order (major premise, then minor premise, then conclusion).
\Cref{fig:example_syllogism} is an example of a standard-form syllogism.

\paragraph{Mood and Figure.}
The \textit{mood} of a categorical syllogism consists of the letter names of the propositions it contains.
For example, the mood for the syllogism presented in \Cref{fig:example_syllogism} is AAA.
The \textit{figure} of a categorical syllogism is determined by the location of the two occurrences of the middle term in the premises. As shown in \Cref{tb:figure_4_types}, there are 4 possible figures.
To accurately determine the mood and figure of a categorical syllogism, it must be in standard form (defined above). 
Any standard-form syllogism is completely described when we specify its mood and figure.
To simplify the terminology, in this paper, we define the combination of mood and figure as the \textit{configuration} of this syllogism.

\paragraph{Valid Inference Types.}
Since there are 4 kinds of categorical propositions and 3 categorical propositions in a categorical syllogism, there are 64 possible moods ($4^3=64$).
As each mood can occur in each of the four figures, in total we have $4^4=256$ different syllogisms. 
Among these, only 24 are valid forms, which are extensively studied by logicians.\footnote{15 configurations are ``unconditionally valid'' and another 9 are ``conditionally valid''. It is related to existential import in \Cref{sec:suggestions}.}
Thus, we have the following fact: \textit{the validity of the standard syllogism can be determined by checking the configuration (mood and figure) against a list of valid syllogistic forms}.

\begin{table}[]
\small
\centering
\begin{tabular}{lcccc}
\toprule
Figure &  1 &  2 &  3 &  4 \\\midrule
Major Premise & \textbf{M} - P & P - \textbf{M} & \textbf{M} - P & P - \textbf{M} \\
Minor Premise & S - \textbf{M} & S - \textbf{M} & \textbf{M} - S  & \textbf{M} - S \\
Conclusion & S - P & S - P & S - P & S - P \\\bottomrule
\end{tabular} 
\caption{Categorical syllogisms have 4 different figures.}
\label{tb:figure_4_types}
\end{table}

\subsection{Analyzing Syllogisms as a Logician}
\label{sec:logician_approach}

We now briefly go through the steps that logicians take for an ordinary categorical syllogism \cite{copi2011introduction, hurley2012concise}.

\paragraph{Translating Categorical Propositions.}
In practice, rare propositions are in their standard form and we need to make translations.
The major benefit of such translation is that the operations and inferences pertinent to standard-form categorical propositions can be directly applied to these statements.
Logicians have developed a number of well-tested methods for translating non-standard propositions, although given the richness of ordinary language, these specific rules can not cover all possible cases. 

\paragraph{Determining the Mood and Figure.}
Once a categorical syllogism is written in its standard form, its figure and mood can be determined by comparing it to \Cref{tb:type_propositions,tb:figure_4_types}.
The judgment of a syllogism's configuration is then rather straightforward.

\paragraph{Checking Validity.}
For a given standard-form categorical syllogism, there are at least the following three ways to check its validity: (1) Use the configuration of this syllogism and then compare it against a list of pre-defined valid syllogistic forms; (2) Use the method of Venn diagrams or generalized quantifier theory to perform set operations; or (3) Check to see if the syllogism conforms to certain rules that are developed by logicians.

\paragraph{Handling Non-Standard Cases.}
When translating into standard-form syllogisms, some specific cases are worth attention, including the treatment of singular propositions, syllogisms with more than three terms, and enthymemes and sorites. We provide the details of these situations in \Cref{appendix:special_cases}.

\begin{table*}[]
\resizebox{0.995\textwidth}{!}{
\begin{tabular}{lllccccllllclrc}
\toprule
\multirow{2}{*}{} & \multicolumn{2}{c}{Data Generation} &  & \multicolumn{4}{c}{Annotation} &  & \multicolumn{3}{c}{Performance} &  & \multicolumn{2}{c}{Meta} \\\cmidrule{2-3} \cmidrule{5-8} \cmidrule{10-12} \cmidrule{14-15} 
 & Method & Source &  & Term & Mood & Figure & Validity &  & Task & Model & Acc. &  & \multicolumn{1}{l}{Total} & Access \\ \midrule
\multicolumn{15}{c}{\textit{\underline{Syllogisms Datasets}}}\\
SylloFigure & \multicolumn{2}{l}{\multirow{2}{*}{Entailment part of SNLI}} &  & \multirow{2}{*}{Middle} & \multirow{2}{*}{\redsign} & \multirow{2}{*}{1-4} & \multirow{2}{*}{Entail} & & Figure  & \multirow{2}{*}{BERT} & \multirow{2}{*}{92\%} &  & \multirow{2}{*}{8,635}  & \multirow{2}{*}{Yes}  \\
\cite{10.1007/978-3-030-81197-6_37} & & &&&& &&& identification \\\midrule
Avicenna & \multirow{2}{*}{Crowdsourcing} & Books, &  & \multirow{2}{*}{Middle} & \multirow{2}{*}{\redsign} & \multirow{2}{*}{\redsign} & \multirow{2}{*}{valid, invalid} &  & Conclusion & GPT-2 trans. & \multirow{2}{*}{32.0\%} &  & \multirow{2}{*}{6,000} & \multirow{2}{*}{Yes}\\
\cite{Aghahadi2022-AGHAAC} & & articles, etc. & & & & & & & generation & learning & \\\midrule
SylloBASE & Template w/  & Wikidata &  & \multirow{2}{*}{\greysign} & \multirow{2}{*}{\greysign} & \multirow{2}{*}{\greysign} & \multirow{2}{*}{valid, invalid} &  & Conclusion & \multirow{2}{*}{RoBERTa} & \multirow{2}{*}{72.8\%} &  & \multirow{2}{*}{51,000} & \multirow{2}{*}{No} \\
\cite{wu-etal-2023-hence} & GPT-3 rewrite & ConceptNet & &&&& & & selection &  \\\midrule
Logical & \multicolumn{2}{l}{\multirow{2}{*}{Human authored questions}} &  & \multirow{2}{*}{\greysign} & \multirow{2}{*}{\redsign} & \multirow{2}{*}{\redsign} & valid &  & Conclusion validity  & \multirow{2}{*}{PaLM 2-L} & $\sim$90\% &  & \multirow{2}{*}{48} & \multirow{2}{*}{No} \\
\cite{dasgupta2023language} &  & & & & & & belief-consistent & & identification 
 & & (support) \\\midrule
NeuBAROCO & \multicolumn{2}{l}{BAROCO (originally designed} &  & \multirow{2}{*}{\redsign} & \multirow{2}{*}{\greysign} & \multirow{2}{*}{\redsign} & entail, contra, neu  &  & Conclusion validity & \multirow{2}{*}{GPT-3.5} & 51.7\% &  & \multirow{2}{*}{375} & \multirow{2}{*}{No} \\
\cite{ando-etal-2023-evaluating} & \multicolumn{2}{l}{for human intell. test)} & & && & inference types & & identification & & (overall) \\\midrule
Reasoning & \multirow{2}{*}{Template} & Hand-crafted &  & \multirow{2}{*}{\yellowsign}  & \multirow{2}{*}{\yellowsign} & \multirow{2}{*}{\yellowsign}  & \multirow{2}{*}{valid, invalid}  &  & Conclusion  & \multirow{2}{*}{PaLM 2} & \multirow{2}{*}{$\sim$75\%} &  & \multirow{2}{*}{1,920} & \multirow{2}{*}{Yes} \\
\cite{eisape2024systematic} &  & triples list &  &&&& & & selection  \\\midrule
\multicolumn{15}{c}{\textit{\underline{First-order Logic Datasets}}}\\
FOLIO & Template w/ crowd & \multirow{2}{*}{N/A} & & \multirow{2}{*}{\greysign} & \multirow{2}{*}{\greysign} & \multirow{2}{*}{\greysign} & true, false, & & Conclusion truth  & Logic-LM & \multirow{2}{*}{78.1\%} & & \multirow{2}{*}{1,435} & \multirow{2}{*}{Yes}\\
\cite{han2022folio} & -sourcing rewrite & & & & & & unknown & & identification & (GPT-4)\\\midrule
ProntoQA & \multirow{2}{*}{Template} & Generated & & \multirow{2}{*}{\yellowsign} & \multirow{2}{*}{\redsign} & \multirow{2}{*}{\redsign} & true, &  & Validity of & \multirow{2}{*}{GPT-3} & \multirow{2}{*}{$\sim$90\%} & & \multirow{2}{*}{400} & \multirow{2}{*}{Yes}\\
\cite{saparov2023language} & & ontology & & &  & & false & & sorites &  \\
\bottomrule
\end{tabular}
}
\caption{Overview of existing syllogism datasets, along with their construction methods, annotations included, and the documented model performance. 
\yellowsign~denotes annotations could be inferred based on the provided dataset construction method, \greysign~denotes annotations are generated in the intermediate steps of the dataset construction but are neither released nor inferred, and \redsign~denotes annotations not available or no information.}
\label{tb:summary_datasets}
\end{table*}

\section{Review of Existing Syllogism Datasets}
\label{sec:existing_datasets_overview}

\subsection{Summary of Syllogism Datasets}
\label{sec:datasets_classify}

We categorize all existing syllogism datasets based on their construction methods, i.e., how the text of premises and conclusions are generated.
In real practice, although some datasets are originally proposed for predicate (first-order) logic, their construction methods might involve syllogisms, or a portion of or the whole dataset contains only categorical propositions.
As these datasets could be formulated as syllogisms, we also list two representative ones for completeness.
All syllogism datasets are summarized in \Cref{tb:summary_datasets}.\footnote{Some prior works use syllogism datasets that are not in the format of natural language, such as \citet{dong2020learning}. We skip the discussions of these studies.}

\paragraph{Template-based Approach.}
Datasets falling into this category are normally generated using templates, i.e., four standard propositions in \Cref{tb:type_propositions}.
The relation triplets are sampled from different sources and then filled into terms positions of these templates to form the complete syllogisms.
For example, questions in ProntoQA \cite{saparov2023language} use ontology generation and contain a series of premises and thus essentially sorites. 
\citet{eisape2024systematic} use a list of 30 relation triplets, the terms of which have no obvious semantic associations.
The relation triplets in \citet{wu-etal-2023-hence} are sampled from Wikidata and ConceptNet, and the propositions generated from templates are further rephrased by using GPT-3.

\paragraph{Text Generated by Humans.}
Non-synthetic datasets are normally developed through crowdsourcing efforts. 
To acquire high-quality inference questions efficiently, these datasets sometimes rely on guidance during the crowdsourcing tasks.
SylloFigure \cite{10.1007/978-3-030-81197-6_37} is built based on the idea of enthymeme reconstruction. Specifically, \citet{10.1007/978-3-030-81197-6_37} select the entailment part of the SNLI \cite{bowman-etal-2015-large} dataset and then add the annotations of figures.
Avicenna \cite{Aghahadi2022-AGHAAC} is a crowdsourcing dataset, and the syllogisms are extracted from various sources, such as books and news articles. 
Syllogisms in \citet{dasgupta2023language} are hand-authored.
NeuBAROCO \cite{ando-etal-2023-evaluating} originates from BAROCO, which is written in Japanese and is developed to evaluate human syllogistic reasoning abilities.
FOLIO \cite{han2022folio} first generates logically valid stories using syllogism templates and then asks human annotators to write logically valid stories in natural language.

\paragraph{Our Newly Collected Test Examples.}
As shown in \Cref{tb:summary_datasets}, nearly all datasets with human-generated text lack certain kinds of annotations, thus causing troubles in analyzing them (in \Cref{sec:coverage}).
We fill in this missing gap by collecting relevant examples and corresponding exercise questions from standard introduction to logic textbooks \cite{copi2011introduction, kelley2013art, baronett2008logic, hurley2012concise}.

In total, we collect 371 examples of translating statements into standard form, covering all the possible forms of phraseology discussed in \Cref{sec:variations_list}; 64 examples for judging the types of standard propositions; and 116 examples for judging the validity of a given syllogism, with complete annotations for the mood and figure. Among these examples, 57 are enthymemes. 

\subsection{Variations of Categorical Syllogisms}
\label{sec:variations_list}

A set of questions that cover all the possible cases could be achieved by varying components of different levels of granularity that we outline in \Cref{sec:cate_syll}.
We consider all possible variations from two angles: syllogisms in standard and non-standard forms. For standard syllogisms, the underlying nature is decided by the combination of mood and figure, which leads to 256 different cases.

For non-standard syllogism, there are variations both on the individual proposition level and the syllogism level.
On the proposition level, we consider the different options of quantifiers, terms, and copula:
(1) Besides standard quantifiers, the propositions could have non-standard quantifiers (also known as generalized quantifiers), such as ``few'', ``a few'', ``not every'', or ``anyone'', and unexpressed quantifiers;
(2) Terms could be expressed with only an adjective, a plural noun or a pronoun, and the verbs are in other forms of the verb ``to be;'' and
(3) Certain propositions could be typically translated into categorical propositions. Established categories include singular propositions, conditional statements such as ``if ... then,'' exclusive propositions that involve words ``only,'' ``none but,'' and ``none except,'' and exceptive propositions in the form of ``All except S are P'' and ``All but S are P''.

On the syllogism level for non-standard syllogism, we vary the following (details in \Cref{appendix:special_cases}): 
(1) It is possible that the syllogism covers more than three terms; and 
(2) Besides the normal syllogisms with two premises and one conclusion, there exist situations with more than two premises or missing premises, which we refer to as enthymemes and sorites.

On top of all the options above, instead of putting the propositions in a well-structured format (i.e., explicitly listing them as premises and conclusions), we could mix them all together as ordinary arguments. 
Some other parts could be varied, such as the order of the two premises. Since the change of the ordering does not change the validity of the conclusion, we skip the discussion of this part.

\subsection{Coverage of Current Datasets}
\label{sec:coverage}

In \Cref{sec:variations_list}, we have enumerated all the possible cases of categorical syllogisms.
In this section, we will use this checklist to evaluate the coverage of current syllogism datasets. We mainly consider the following aspects: (1) the forms of phraseology covered, and (2) the mood and figure covered in these syllogism datasets.

\subsubsection{Building Tools for Assessing Coverage}
\label{sec:tools}

Most of the datasets do not have the annotations needing to be assessed (details in \Cref{tb:summary_datasets}).
Making up these missing pieces would require human annotators with linguistic background.
Given the huge amount of human effort for such annotations, we take the approach of directly asking LLMs to perform as an annotator for labeling.

To ensure that we can build prompts with reasonable performance, we calibrate them on our newly collected textbook questions (discussed in \Cref{sec:datasets_classify}).
We also use the fact about the validity of syllogisms mentioned in \Cref{sec:cate_syll} for cross-checking: for a valid inference, if a predicted configuration is not one of the valid syllogism forms, then there is something wrong with this prediction.

\paragraph{Translating the Propositions.}
When translating statements, besides a deep understanding of the given statement, we need to follow some established rules set by logicians (for example, the treatment of singular propositions discussed in \Cref{appendix:special_cases}).
We thus base our prompt design on a 2-step translation process: (1) determine the nature of a proposition by classifying it into categories listed in \Cref{tb:prop_category}), and (2) then perform the translation based on the set rules within that category. To make sure the translated proposition is in the standard form, we also set up a mechanism for a second-round translation.
We observe that GPT-4o performs well in identifying the forms of phraseology, while it is easy to incorrectly classify some statements into singular propositions.
A manual check for the translated propositions shows that GPT-4 achieves 87.3\% accuracy on 371 textbook problems, with 68 propositions translated twice.

\paragraph{Judging the Mood and Figure.}
We can not first translate individual propositions and then simply combine the detected proposition types together to form the mood of the syllogism, due to the issue of having potentially more than three terms 
 (in \Cref{appendix:special_cases}).
Thus, we feed the syllogism as a whole and ask GPT-4 to generate the mood and figure simultaneously. The principles and rules discussed above for translating propositions are also incorporated into the prompt.
Experimental results on 116 textbook examples reveal an accuracy of 87.9\% for mood detection, 48.3\% for figure detection, and 44.8\% for configuration detection.
A further review of mood detection results reveals that this high accuracy is due to the fact that most of the collected textbook examples are standard-form propositions.

\subsubsection{Datasets Coverage Observations}
\label{sec:coverage_observations}

\begin{table}[]
\small
\resizebox{0.49\textwidth}{!}{
\begin{tabular}{llrrr}
\toprule
 &  & \multicolumn{1}{l}{SylloFigure} & \multicolumn{1}{l}{Avicenna} & \multicolumn{1}{l}{Reasoning} \\\midrule
\multirow{6.5}{*}{Proposition} & Standard (\%) & 0.9 & 0.6 & 100 \\
 & Singular (\%) & 64.7 & 27.2 & 0 \\
 & Condition (\%) & 2.3 & 9.5 & 0 \\
 & Exclusive (\%) & 0.1 & 1.0 & 0 \\
 & Others (\%) & 32.0 & 61.7 & 0 \\\cmidrule{2-5}
 & Total & 2,448 & 1,864 & 2,560 \\
 \midrule
\multirow{4}{*}{Configuration} & Coverage (\%) &  $>$4.3 & $>$2.7 & 100 \\
 & Actual count & $>$11 & $>$7 & 256 \\\cmidrule{2-5}
& Syllo assessed (\%) & 71.1 & 60.9 & 100 \\\midrule
\multicolumn{2}{c}{Total syllogisms} & 868 & 622 & 2,560 \\
\bottomrule
\end{tabular}
}
\caption{Forms of phraseology and configurations of categorical syllogisms covered in datasets.}
\label{tb:prop_category}
\end{table}

We apply our calculating tools developed in \Cref{sec:tools} to all three categorical syllogism datasets currently released. For the SylloFigure and Avicenna datasets, we conduct analyses only on the test sets, while for the Reasoning dataset, we randomly sample 10 relation triples out of 30 and then generate the complete syllogisms.
We use the whole dataset for assessing the proposition forms, since it is rather straightforward.
Regarding the underlying configuration of syllogisms: As the Reasoning dataset is generated by using templates, the whole dataset could be accurately assessed (since we have all the annotations such as mood and figure).
Using the cross-checking method discussed in \Cref{sec:tools}, we estimate 60.9\% of syllogisms could be properly assessed in the Avicenna dataset, while a higher 71.1\% for the SylloFigure dataset, as it contains human annotated figures.

Our assessment results are reported in \Cref{tb:prop_category} (detailed configurations are in \Cref{fig:gpt_4-256_cases}). In \Cref{fig:prop_correct_wrong} we also provide the distribution of the estimated proposition types in the SylloFigure dataset.
We observe that both the proposition types (A, E, I, O) and forms of phraseology (types in \Cref{tb:prop_category}) are distributed highly unevenly, and datasets normally have different distributions. 
Regarding the coverage of configurations, we observe that compared to template-based datasets, datasets using human-generated text are normally centered on a few specific moods and figures, i.e., Avicenna covers only over 7 different syllogisms configurations, calculated from 60.9\% of the whole dataset.

Since we use LLMs instead of human effort to make up the missing mood and figures, the coverage percentages in \Cref{tb:prop_category} can only be treated as rough estimates.
Nevertheless, our key point is clear: datasets that are from crowdsourcing efforts are skewed to certain linguistic styles and cover only limited configurations of syllogisms.
We thus suggest researchers take the actual variations covered by the datasets into account when interpreting experimental results. 

\section{Evaluating LLMs for Analyzing Syllogisms}
\label{sec:exp}

\subsection{What Do We Know So Far?}

\paragraph{Reported Results for Validity Inferences.}
We observe prior studies mainly make use of the following approaches to evaluate the validity of categorical syllogisms: (1) given two premises, select a correct conclusion from multiple choices \cite{wu-etal-2023-hence, eisape2024systematic}, (2) given two premises and a conclusion, identify if the logic inference is valid \cite{dasgupta2023language, ando-etal-2023-evaluating}, and (3) given two premises or more, generate the conclusion \cite{Aghahadi2022-AGHAAC, saparov2023language}.
In general, most prior works report LLMs have an accuracy of around 75\% when evaluating the validity of given syllogisms. 
We provide more performance evaluation details in \Cref{tb:summary_datasets}.

\paragraph{Error Analysis.}
One trend for analyzing the errors that LLMs make is to compare them with human cognition biases.
\citet{dasgupta2023language} find that like humans, LLMs give out more accurate answers when the semantic content of a task supports the logical inferences.
\citet{ando-etal-2023-evaluating} analyze the models' errors from three aspects: belief biases, conversion errors, and atmosphere effects.
\citet{eisape2024systematic} provide more direct observations that LLMs replicate some human biases discovered in psychology studies, while LLMs could overcome these biases in certain situations. 

\subsection{LLMs' Performance Breakdowns by Syllogisms Configurations}

In this section, we reproduce the experimental results of LLMs for judging the logical validity of syllogisms and check to see if prior findings still hold.
We will also break down the error rate by the configurations of syllogisms.

\subsubsection{Setups}

\paragraph{Models and Datasets.}
We conduct our experiments using OpenAI's GPT models, as they are commonly used large language models with compelling performance on a variety of inference tasks \cite{openai2024gpt4}.
All our experiments are done using GPT-4 and GPT-4o.
We use the same set of datasets that we assess in \Cref{sec:coverage}. The details of these datasets are provided in \Cref{sec:coverage_observations}.

\paragraph{Prompts Used.}
For comparison purposes, we follow the chain-of-thought prompt used in \citet{eisape2024systematic} and test how LLMs perform logical inferences under a zero-shot learning setting. 

\begin{table}[]
\small
\centering
\resizebox{0.32\textwidth}{!}{
\begin{tabular}{lrcc}
\toprule
Dataset & \#  & GPT-4 & GPT-4o \\\midrule
SylloFigure & 868 & 74.3 & 70.2 \\
Avicenna & 622 & 72.5 & 53.4 \\
Reasoning & 2,560 & 90.2 & 95.4 \\
\bottomrule
\end{tabular}
}
\caption{Accuracy (\%) for checking the validity of categorical syllogisms.}
\label{tb:blocks_perf}
\end{table}

\begin{figure}[!ht]
\centering
%% first line
\subfigure[Reasoning dataset (GPT-4)]{
\begin{minipage}[h]{0.99\linewidth}
\centering
\includegraphics[width=0.99\linewidth]{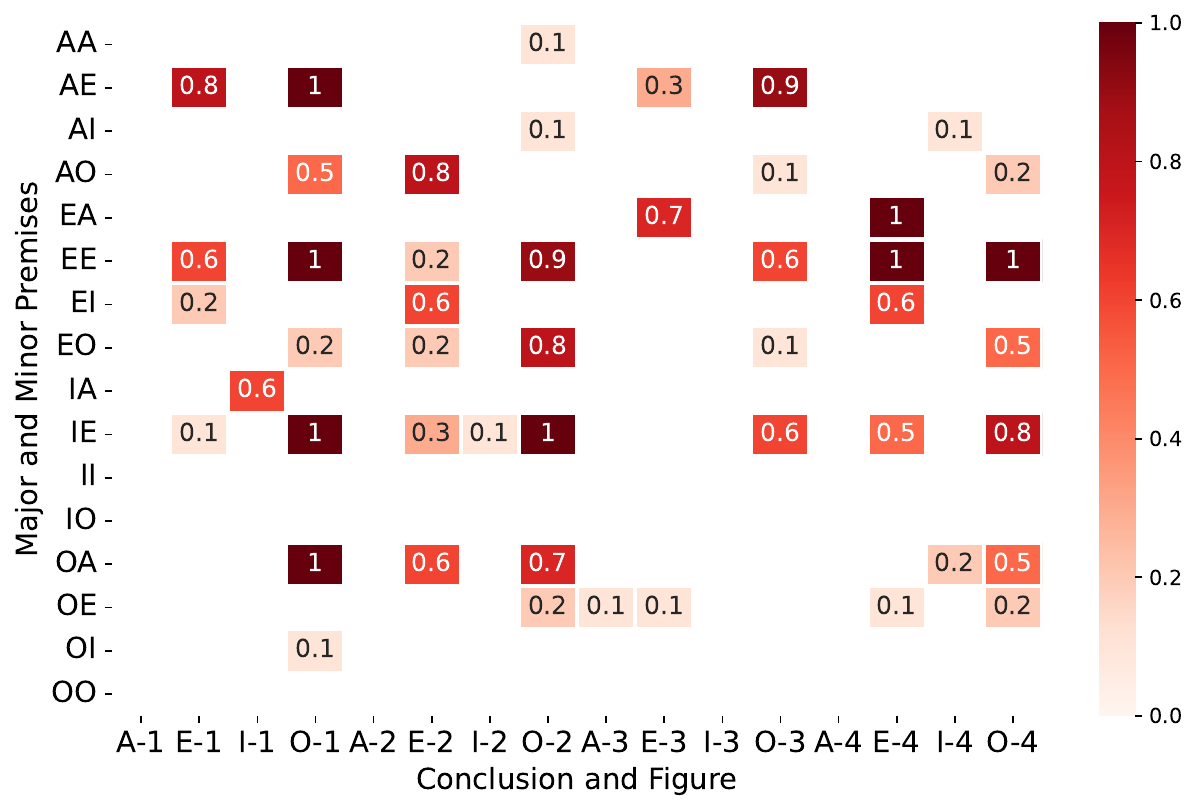}
%\caption{fig2}
\end{minipage}%
\label{fig:error_rate_1}
}%
\\
\subfigure[Reasoning dataset (GPT-4o)]{
\begin{minipage}[h]{0.99\linewidth}
\centering
\includegraphics[width=0.99\linewidth]{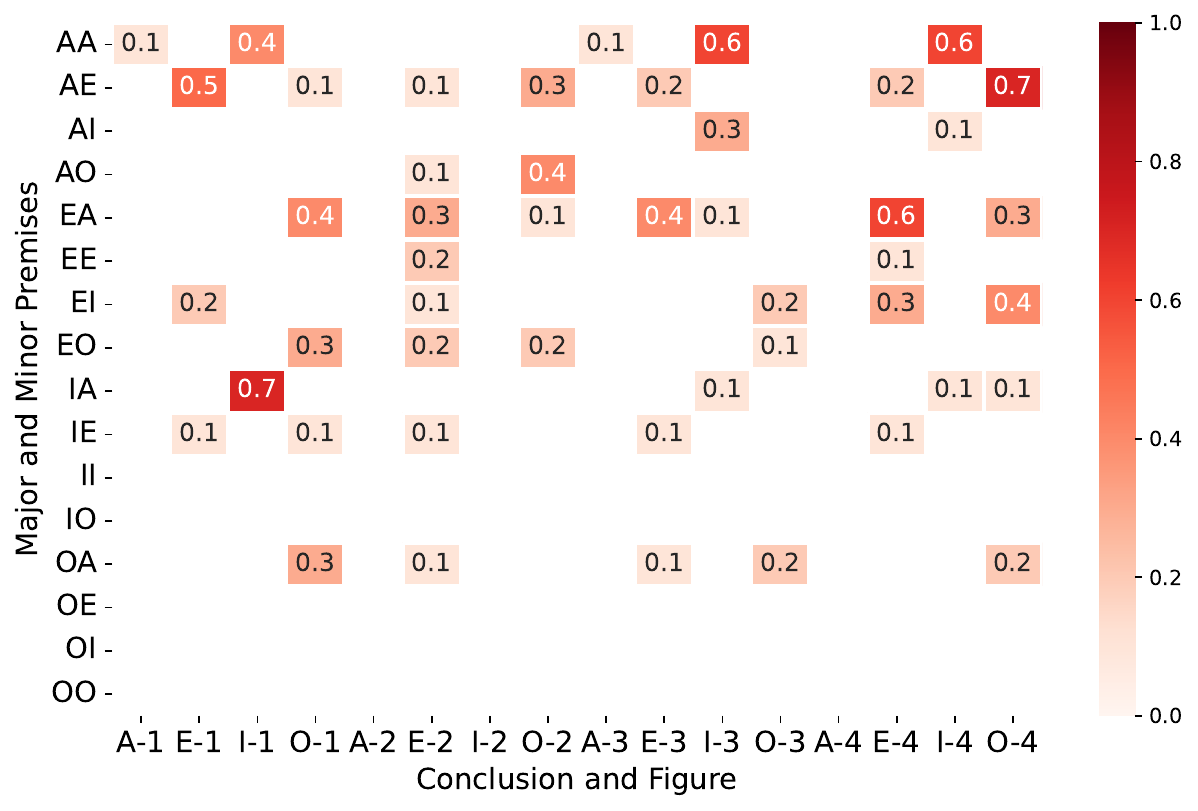}
%\caption{fig2}
\end{minipage}
\label{fig:error_rate_2}
}
\subfigure[SylloFigure and Avicenna datasets]{
\begin{minipage}[h]{0.99\linewidth}
\resizebox{0.99\textwidth}{!}{
\begin{tabular}{clrrrllrrr}
\toprule
\multicolumn{1}{l}{\multirow{2.5}{*}{Figure}} & \multicolumn{4}{c}{SylloFigure} &  & \multicolumn{4}{c}{Avicenna} \\ \cmidrule{2-5} \cmidrule{7-10} 
\multicolumn{1}{l}{} & Mood & \multicolumn{1}{l}{\#} & \multicolumn{1}{l}{GPT-4} & \multicolumn{1}{l}{GPT-4o} &  & Mood & \multicolumn{1}{l}{\#} & \multicolumn{1}{l}{GPT-4} & \multicolumn{1}{l}{GPT-4o} \\\midrule
\multirow{4}{*}{1} & AAA & 47 & 0.21 & 0.28 &  & AAA & 310 & 0.20 & 0.42 \\
 & AAI & 38 & 0.32 & 0.42 &  & AAI & 12 & 0.33 & 0.42 \\
 & AII & 502 & 0.21 & 0.26 &  & AII & 25 & 0.28 & 0.68 \\
 & N/A & 56 & 0.34 & 0.32 &  & EAE & 2 & 1 & 0.50 \\\midrule
\multirow{2}{*}{2} & EAE & 1 & 0 & 0 &  & EAE & 3 & 0 & 0 \\
 & N/A & 180 & 0.28 & 0.36 &  & AEE & 3 & 0.67 & 0.33 \\\midrule
\multirow{3}{*}{3} & AAI & 2 & 1 & 0.5 &  & AAI & 1 & 0 & 0 \\
 & AII & 26 & 0.54 & 0.38 &  & AII & 4 & 0 & 0 \\
 & N/A & 8 & 0.38 & 0.38 &  & IAI & 2 & 0.50 & 0.50 \\\midrule
\multirow{2}{*}{4} & IAI & 1 & 1 & 0 &  & IAI & 14 & 0.29 & 0.64 \\
 & N/A & 7 & 0.71 & 0.43 &  & AAI & 3 & 0 & 0.33 \\\midrule
N/A & \multicolumn{4}{c}{/} & & N/A & 243 & 0.35 & 0.52 \\
\bottomrule
\vspace{.01cm}
\end{tabular}
}
\end{minipage}
\label{fig:error_rate_3}
}
\caption{Error rate ($\downarrow$) of GPT-4 and GPT-4o using zero-shot chain-of-thoughts. 
(a) and (b): Breakdowns on all 256 configurations of categorical syllogisms in the Reasoning dataset, calculated over 10 different combinations. 
A white block indicates an error rate of 0 (thus 100\% accuracy) in that specific configuration. (c): Breakdowns by configurations in the SylloFigure and Avicenna datasets. We mark the predicted configuration as ``N/A'' if it does not pass the cross-check discussed in \Cref{sec:tools}.}
\label{fig:gpt_4-256_cases}
\end{figure}

\subsubsection{Results}

We visualize the error rate of GPT-4 and GPT-4o on the complete 256 configurations of syllogisms from the Reasoning dataset in \Cref{fig:error_rate_1,fig:error_rate_2}.
The error rate in the SylloFigure and Avicenna datasets are reported in \Cref{fig:error_rate_3}.
We also report the total accuracy of validity judgment in \Cref{tb:blocks_perf} for reference purposes.

We observe the following trends.
(1) Comparing \Cref{fig:error_rate_1} with \Cref{fig:error_rate_2}, we observe different patterns for the configurations of syllogism that LLMs fail. For example, GPT-4 nearly has no errors when two premises are in AA format, while GPT-4o makes even more than half of the mistakes for AAI-3 and AAI-4. However, GPT-4o performs better than GPT-4 for configurations that GPT-4 has 0\% accuracy.
(2) For two datasets with human-written text, GPT-4 seems to have more stable performance compared to GPT-4o, i.e., the error rate in \Cref{fig:error_rate_3} is roughly the same for AAA-1, AAI-1, and AII-1.
(3) We observe that for the same configuration, LLMs generally have a higher error rate in human-generated SylloFigure and Avicenna datasets (\Cref{fig:error_rate_3}), compared to the template-based Reasoning dataset (\Cref{fig:error_rate_1,fig:error_rate_2}).
It seems to suggest that translating the syllogisms to the standard form is the bottleneck for LLMs to behave well, as the only difference that the Reasoning dataset has is the expressed way of the premises and conclusions. The underlying ability required to infer remains unchanged: if LLMs can translate ordinary text into the standard format, then it should work well.
This observation also aligns with the challenges of the logicians' approach for analyzing syllogisms: as discussed in \Cref{sec:cate_syll}, the most difficult part is translating the propositions -- once the mood and the figure are determined, then checking the validity of the syllogism is trivial. 

\begin{figure}
\centering
\includegraphics[width=0.99\linewidth]{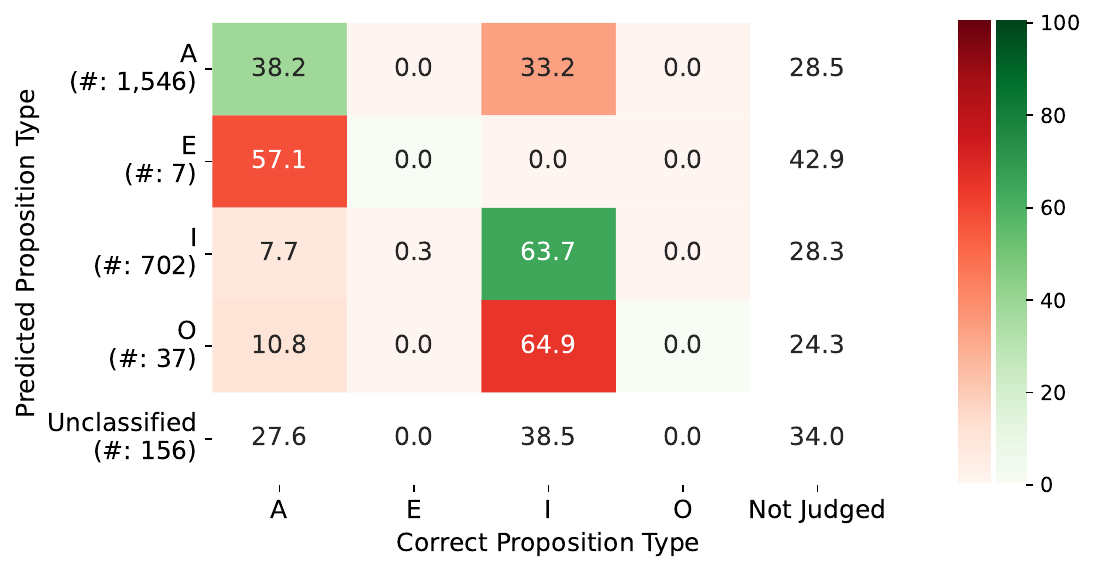}
\caption{Percentage breakdowns of the correct propositions within each predicted proposition type (by GPT-4). 
156 propositions (last row) could not be classified and we can not automatically verify the correctness of predictions without human efforts (last column).}
\label{fig:prop_correct_wrong}
\end{figure}

\subsection{Ambiguity of Natural Language}
\label{sec:bottleneck}

Our observation is that translating into standard propositions is the most challenging part for LLMs and thus causes errors. 
In this section, we take a closer look at the types of translation errors LLMs make, especially around quantifiers.

We visualize in \Cref{fig:prop_correct_wrong} the percentage of correct propositions within each predicted proposition type in the SylloFigure dataset. 
We observe that in general, the LLMs have a much higher accuracy in recognizing the ``some'' quantifier, although it sometimes mixes the particular negative type (O) with the particular affirmative type (I).
We also observe LLMs tend to confuse universal affirmative (A) with particular affirmative (I): 
among 1,546 propositions that are predicted as universal affirmative type (A), 33.2\% should be particular affirmative (I). 
This phenomenon is related to the interpretation of singular propositions (discussed in \Cref{appendix:special_cases}) and is also partially due to the fact that singular propositions represent a huge portion of the SylloFigure dataset (shown in \Cref{tb:prop_category}).

We shall point out that analyzing the sensitivity of quantifiers by LLMs is not entirely new in computation linguistics. 
One representative work is \citet{cui-etal-2022-generalized-quantifiers}, where the authors rely on generalized quantifier theory to quantify their contribution to the errors of NLU models. 
There is a recent work by \citet{madusanka-etal-2023-quantifiers} that tries to investigate how different generalized quantifiers affect LLMs by employing a textual entailment problem.
\citet{ando-etal-2023-evaluating} also suggest the importance of differentiating the problems of interpreting quantifiers and negations from performing logical inferences.
In this work, we hold the same standpoints that the comprehension of quantifiers greatly affects the model performance and future models should enhance their abilities to analyze quantifiers.
Compared to these prior studies, we present a more complete and comprehensive analysis of quantifiers in a specific syllogism setting. 

\section{Moving Forward: Future Directions}
\label{sec:discussion}

\subsection{Suggestions for Future Datasets}
\label{sec:suggestions}

\paragraph{Existential Import.}
In \Cref{sec:cate_syll}, we mention that there are 24 valid configurations over all 256 cases, 9 of which rely on the existential import assumption.
We notice that nearly all prior works, except \citet{ando-etal-2023-evaluating}, implicitly make such an assumption.  
We recommend researchers explicitly mention this assumption in their dataset release, as it affects the determination of the validity of syllogisms \cite{hurley2012concise}.

\paragraph{Complete Annotations.}
As shown in \Cref{tb:summary_datasets}, many syllogism datasets lack certain kinds of annotations, thus causing trouble when we try to assess the coverage of language variations in \Cref{sec:coverage}.
We notice that in their dataset descriptions, especially datasets that make use of templates, many annotations are actually generated during the dataset construction process (for example, blocks marked with \greysign~in \Cref{tb:summary_datasets}).
We suggest researchers consider releasing these annotations from intermediate steps to promote a more accurate assessment of the properties of their datasets.

\paragraph{Ordinary Argument.}
We observe that all syllogism datasets in \Cref{sec:coverage} are in a well-structured format, i.e., the premises and conclusions are listed separately.
In real life, however, a more realistic situation is that the premises and conclusions are mixed together, with no clear indications or separators. There might even be cases such as enthymemes.
Thus, one possible direction is to build datasets that contain ordinary arguments.
Building such a dataset will also enable a variety of downstream applications, for example, to evaluate the syllogisms hidden in human forecasts or debates. We note there has been some exploration work in this direction \cite{10.1145/3594536.3595170}.

\subsection{Enhancing Logical Reasoning Abilities}

In prior studies, we observe two lines of research that attempt to enhance the logical reasoning abilities of the LLMs.
One line of approach is to rely on external modules.
\citet{olausson-etal-2023-linc} make use of an external theorem prover, which symbolically performs deductive inference.
\citet{poesia2023certified} propose to augment the LLM's reasoning ability by using externally certified reasoning, such as a theorem-proving environment for incremental proof generation.
Another line is to directly incorporate the reasoning ability inside the LLMs. Representative work includes \citet{xu2024faithful}, which argues that the reasoning ability should be inherited without using any external blocks. In general, it is unclear which type of approach is better.
Specific to our syllogism inference case, if our ultimate goal is to build a trustworthy and reliable system with no tolerance for errors, then enabling some external pure logical solvers would help ensure the accuracy of analyzing syllogisms.

\section{Conclusion}

This work tries to address the question of whether current proposed benchmarks can evaluate logical reasoning abilities accurately and thoroughly. 
We choose categorical syllogism as our main focus, since this logical system has been extensively studied by logicians and has many nice properties, such as a finite number of possible cases, and automated ways of solving it. A categorical syllogism is also arguably the most basic building block for any other more complex reasonings.
We draw the inspirations from how logicians analyze categorical syllogisms and construct a list of variations that should be covered by benchmarks. Our results show that there is no single dataset that properly covers all possible situations.
We also summarize the current progress made in judging the validity of the categorical syllogisms. Our findings highlight the importance of correctly interpreting different quantifiers.
Finally, we provide a discussion of several points that might be worth considering when researchers plan on the future release of categorical syllogism datasets. 

\section*{Limitations}

In this work, we mainly focus on analyzing the existing benchmarks of categorical syllogisms.
Among 6 syllogism datasets listed in \Cref{tb:summary_datasets}, we are only able to assess 3, as others are not publicly released.
Also, we use GPT-4 as an annotation tool instead of human annotators to generate the missing annotations, such as mood, figure, and forms of phraseology.
Although we have taken steps to control the quality of these annotations (as discussed in \Cref{sec:tools}), it is inevitable that there are errors.

\end{CJK*}

\bibliography{custom}

\appendix

\section{Handling Special Cases When Analyzing Categorical Syllogisms}
\label{appendix:special_cases}

\paragraph{Singular Propositions.}
A singular proposition is defined as making a particular individual or object (for example, a specific person, thing, time, or place) belong to a given class. 
Although it is arguable about the treatment of these singular propositions, logicians seem to agree that in general, these propositions are generally converted into universal propositions.

\paragraph{Reducing the Number of Terms.}
A valid syllogism must have exactly three terms.
When more than three terms seem to be involved in an argument of apparently syllogistic form, we may need to reduce the number of terms to three, by either eliminating synonyms or eliminating class components \cite{copi2011introduction}.

\paragraph{Enthymemes and Sorites.}

In real life, we normally do not make explicit mention of all the premises required to support a given conclusion, especially when the premises are obvious or noncontroversial. 
A syllogism with an unstated premise is called an enthymeme \cite{kelley2013art}.
Sorites are defined as a chain of categorical syllogisms in which the intermediate conclusions have been left out \cite{hurley2012concise}.
The standard treatment for analyzing sorites is to first make their intermediate conclusions or steps explicit, then test the validity of obtained syllogisms separately.

\end{document}